\documentclass{article}

\usepackage{PRIMEarxiv}
\usepackage{authblk}
\usepackage{float}  
\usepackage{tabularx}  

\usepackage{makecell}

\usepackage[utf8]{inputenc} 
\usepackage[T1]{fontenc}    
\usepackage{hyperref}       
\usepackage{url}            
\usepackage{booktabs}       
\usepackage{amsfonts}       
\usepackage{nicefrac}       
\usepackage{microtype}      
\usepackage{lipsum}
\usepackage{fancyhdr}       
\usepackage{graphicx}       
\graphicspath{{media/}}     
\usepackage{comment}
\usepackage[numbers]{natbib}
\title{Vision-Language Modeling in PET/CT for Visual Grounding of Positive Findings

}

\author{
  Zachary Huemann\textsuperscript{1}, Samuel Church\textsuperscript{1}, Joshua D. Warner\textsuperscript{1,2}, Daniel Tran\textsuperscript{1}, Xin Tie\textsuperscript{1},
  Alan B McMillan\textsuperscript{1}, Junjie Hu\textsuperscript{1}, Steve Y. Cho\textsuperscript{1,2,3}, Meghan Lubner\textsuperscript{1,2}, Tyler J. Bradshaw\textsuperscript{1} \\
  \textsuperscript{1}University of Wisconsin-Madison, Madison, WI 
    53792 \\
    \textsuperscript{2}University of Wisconsin Health, Madison, WI 53792 \\
    \textsuperscript{3} Carbone Cancer Center, Madison, WI 53792 \\
  \texttt{Corresponding Author: zhuemann@wisc.edu}
}

\begin{document}
\maketitle

\begin{abstract}
Vision-language models can connect the text description of an object to its specific location in an image through visual grounding. This has potential applications in enhanced radiology reporting. However, these models require large annotated image-text datasets, which are lacking for PET/CT. We developed an automated pipeline to generate weak labels linking PET/CT report descriptions to their image locations and used it to train a 3D vision-language visual grounding model.
Our pipeline identifies sentences describing positive findings in PET/CT reports by searching for mentions of standardized uptake values (SUVmax) and axial slice numbers. Initial lesion segmentations are created on the PET images, which are matched with the noted SUVmax and axial slice. Each matched lesion is then refined with an iterative thresholding algorithm to produce the final label. From 25,578 PET/CT exams, we extracted 11,356 sentence-label pairs. Using this data, we trained ConTEXTual Net 3D, which integrates text embeddings from a large language model with a 3D nnU-Net via token-level cross-attention. The model’s performance was evaluated on 251 radiologist-reviewed cases and compared against LLMSeg, a 2.5D version of ConTEXTual Net, and two nuclear medicine physicians. We analyzed performance across different language models, dataset sizes, tracer types, and lesion characteristics. We evaluated detection performance using F1 score, where a predicted contour was considered a true positive if its SUVmax matched the ground truth SUVmax within 0.1 SUV.
The weak-labeling pipeline accurately identified lesion locations in 98\% of cases (246/251), with 7.5\% requiring boundary adjustments. ConTEXTual Net 3D achieved an F1 score of 0.80, outperforming LLMSeg (F1=0.22) and the 2.5D model (F1=0.53), though it underperformed both physicians (F1=0.94 and 0.91). The model achieved better performance on FDG (F1=0.78) and DCFPyL (F1=0.75) exams, while performance dropped on DOTATE (F1=0.58) and Fluciclovine (F1=0.66). The model performed consistently across lesion sizes but showed reduced accuracy on lesions with low uptake.
Our novel weak labeling pipeline accurately produced an annotated dataset of PET/CT image-text pairs, facilitating the development of 3D visual grounding models. ConTEXTual Net 3D significantly outperformed other models but fell short of the performance of nuclear medicine physicians. Our study suggests that even larger datasets may be needed to close this performance gap.

\end{abstract}

\keywords{ Large Language Models, Vision-Language Models, Segmentation, PET/CT, Interactive Reporting
}

\section{Introduction}
Vision-language models (VLMs) have demonstrated impressive capabilities in a range of multimodal tasks, many of which have direct applications in medicine \cite{li2023llavamedtraininglargelanguageandvision, pellegrini2023radialoglargevisionlanguagemodel}. One such task is visual grounding, where the model links text descriptions to specific regions in medical images. A particularly promising application of visual grounding is the creation of interactive radiology reports, which visually map physician-reported findings, such as 'hypermetabolic lymph nodes,' onto the corresponding anatomical locations. This approach can improve communication by making complex imaging data more accessible to patients and physicians alike \cite{folio_multimedia-enhanced_2018}. Beyond enriched communication, visual grounding can assist trainees in connecting clinical descriptions to imaging findings, streamline longitudinal reads by linking prior descriptions to specific image findings, and enhance efficiency through real-time segmentation and automated measurements. Additionally, these models can aid in error detection by providing real-time visualizations of textual descriptions, highlighting discrepancies such as laterality and word substitution errors \cite{lee_detection_2015, vosshenrich_revealing_2021}.

Despite the potential impact of visual grounding in radiology, prior work in this space has been limited. In 2D radiography, vision-language models have used text descriptions to improve segmentation accuracy compared to vision-only models \cite{huemann_contextual_2024, zhao_cap2seg_2024, 10635189}. Maira-2 goes a step further and generates reports by integrating 2D grounded findings \cite{bannur2024maira2groundedradiologyreport}. However, for 3D imaging, few studies have investigated using vision-language models, with most efforts focused on the domain of radiation therapy, where contours are readily available as part of the clinical workflow \cite{rajendran_autodelineation_2025, oh_llm-driven_2024}. However, no prior work has addressed visual grounding in whole-body 3D imaging.  A key challenge is that visual grounding models need very large annotated multimodal datasets, which do not exist for PET/CT imaging. A visual grounding dataset would require linking specific text-defined targets (e.g., 'within the left anterior liver at approximate segment 2A') to corresponding image regions, which is not achieved with whole-report-whole-image alignment methods. This level of granularity presents a significant challenge, making the creation of a dataset for whole-body, disease-agnostic visual grounding an unmet need in nuclear medicine. 

In this work, we develop an automated labeling pipeline that links positive findings in PET/CT reports and their corresponding image locations to create a weakly-labeled visual grounding training dataset for PET/CT. We then use this dataset to train and validate a 3D vision-language visual grounding model for PET/CT. Additionally, we conduct reader studies to evaluate the accuracy of the model.

\section{Methods}
\label{sec:Methods}

\subsection{Data Collection}
Under an institutional review board-approved retrospective protocol with waiver of informed consent, we pulled all whole-body PET/CT exams with radiotracer types [18F]-fluorodeoxyglucose (FDG), DOTATATE, [18F]Fluciclovine, and [18F]DCFPyL conducted from 2010 to 2023 from the university’s picture archiving and communication system (PACS). This resulted in 25,578 PET/CT images and reports.

\subsection{Automated Labeling Pipeline}

\begin{figure*}[!h]
\centerline{\includegraphics[width=0.9\textwidth]{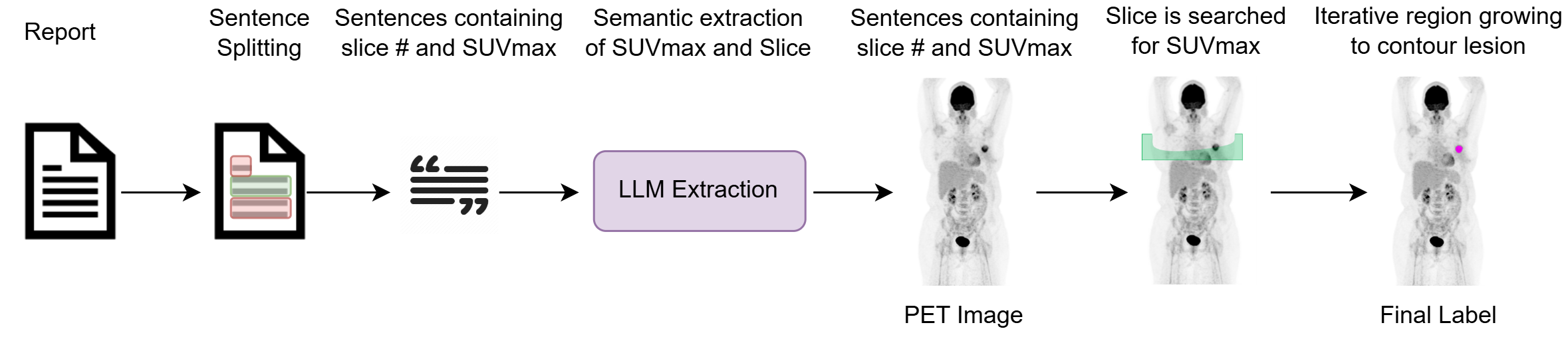}}
\caption{The data preprocessing pipeline is shown above. First, we split the PET reports into individual sentences. We extract all sentences that contain a slice number and an SUVmax. Of those, we check which ones contain anatomical descriptor terms using RadGraph. We then used LLMs and in-context learning to filter out sentences describing prior imaging and sentences containing multiple findings.  For imaging annotations, the reported slice number is searched for the specified SUVmax. If the SUVmax is found, we use an iterative thresholding method to create the label. This results in a training dataset of PET/CT images, descriptive sentences, and referring segmentations for developing a visual grounding model. }

\label{Model}
\end{figure*}

To create the labeled dataset, we use a multi-step process to identify phrases that describe disease findings within the image and then algorithmically generate the label. We utilized the fact that when describing lesions in PET/CT reports, physicians often include the axial slice number and the maximum standardized uptake value (SUVmax) of the noted lesion. We used this SUVmax and slice number information to algorithmically generate an image label, and then redacted the SUVmax and slice number information from the text and used it as the referring expression that describes the specific lesion.

We use a pattern-matching approach to extract sentences containing an SUVmax value and a slice number. This method identifies sentences with the term "SUVmax" followed by a decimal value and the word "slice" followed by an integer. Sentences describing multiple findings across different slices were excluded. RadGraph \cite{jain2021radgraphextractingclinicalentities}, a language model designed to identify entities in radiology reports, was used to detect and discard sentences that contained no anatomical details. This ensured that the model had sufficient context to locate the lesion. 

After identifying candidate sentences containing at least one SUVmax value and slice number, we extracted the current exam’s SUVmax and slice values using in-context few-shot prompting with large language models (LLMs). This approach allowed for semantic value extraction, addressing edge cases that simple rule-based methods often fail to handle; the exact prompt is found in Figure \ref{prompt}. For example, physicians frequently mention previous SUVmax values and slice values alongside current values, making it challenging to accurately extract the relevant information using rule-based methods. To improve the reliability of the LLM-extracted SUVmax and slice, we found it beneficial to ensemble three models for value extraction, accepting a value if at least two models agree. The models used for this process were Mistral-7B-Instruct \cite{jiang2023mistral7b}, Mixstral-8x7B-Instruct \cite{jiang2023mistral7b}, and Dolphin-Instruct \cite{mukherjee2023orca}. Sentences with SUVmax values below 2.5 or templated sentences like those referencing background liver or blood pool were excluded to retain only relevant positive PET findings. 

Once the SUVmax, slice number, and text description were extracted, we then generated corresponding image labels. To achieve this, we applied an iterative image thresholding algorithm \cite{jentzen_improved_2015}, which has been previously validated \cite{weisman_comparison_2020}. The first step of the iterative thresholding method was applied to the 3D PET volume using the extracted SUVmax value to dynamically determine the threshold. The resulting thresholded voxels were then grouped into connected components based on spatial adjacency. For each connected component, the maximum SUV value was calculated. A connected component was selected if its maximum SUV matched the extracted SUVmax within a tolerance of ±0.1 SUV and if the component intersected the specified axial slice.

In most cases, only one connected component met these criteria. Once the correct connected component had been identified, we refined the segmentation through the iterative threshold method \cite{jentzen_improved_2015}. The final result of this pipeline was a 3D contour of a PET-positive finding and its corresponding description. Table \ref{table:data_processing} contains the number of sentences remaining after each preprocessing step. The final 11356 descriptions are from 5126 unique exams with a make of 4798 FDG, 178 DOTATATE, 106 [18F]Fluciclovine, and 44 [18F]DCFPyL.

\begin{table}[H]
\centering
\small
\caption{Data Processing Steps}
\label{table:data_processing}
\resizebox{10cm}{!}{%
\begin{tabular}{lr}
\toprule
Data processing Step & Number of remaining sentences \\[5pt]

\midrule
Starting sentences                                          & 957169  \\
Sentences with SUVmax and slice                             & 50821   \\
Single slice number requirement                             & 43141   \\
Sentences with anatomical description                       & 39758   \\
Sentences with extracted SUVmax and slice                   & 38580   \\
Sentences above 2.5 SUVmax                                    & 32885   \\
Sentences with located SUVmax intersecting slice            & 12484   \\
Sentences with unique SUVmax intersecting slice             & 11356   \\
Sentences with Labels                                         & 11356   \\
\bottomrule
\end{tabular}%
}
\end{table}

The resulting data was then preprocessed for use in model training. The PET, CT, and label were resampled to a resolution of 3 mm. The images were cropped at the top of the skull and subsequently center-cropped or padded to dimensions of 192 × 192 in the axial plane and 352 voxels in length. The sentences had the axial slice and SUVmax information removed, and these were then used as the referring expression in the visual grounding task.

\subsection{Model Architecture}

\begin{figure*}[!h]
\centerline{\includegraphics[width=0.9\textwidth]{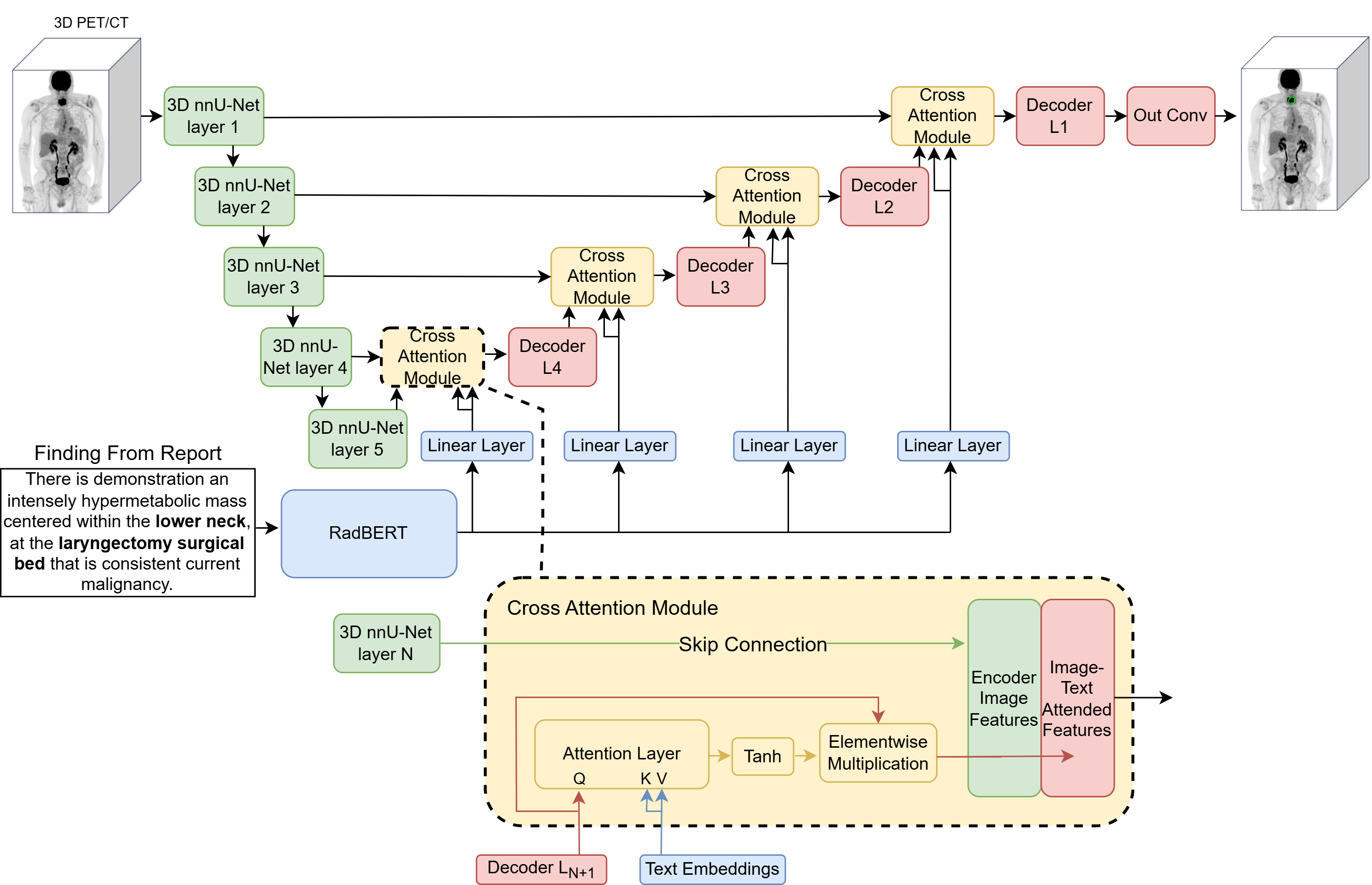}}
\caption{The 3D multimodal vision-language model used for visual grounding of PET findings. The 3D PET/CT is encoded via a 3D nn-Unet and the sentence is encoded via RadBERT. For cross-attention, the language embeddings are used as the key and the value with the vision features as the query, which produces voxel-wise attention maps which are then applied to the voxel space. This mechanism allows for text-guided segmentation.}

\label{Labeling Pipeline}
\end{figure*} 

The model architecture builds on the design principles of ConTEXTual Net, with modifications to incorporate a 3D vision encoder based on the nnU-Net framework \cite{isensee_nnu-net_2021}. The 3D nnU-Net serves as the vision encoder, effectively extracting spatial features across three dimensions simultaneously to enhance segmentation performance. In parallel, a pre-trained language model is used to extract the token-level embedding from the referring expression. These embeddings, projected via a linear layer, provide semantic guidance by encoding anatomical information described in the text. The cross-attention modules, positioned on the decoding side of the nnU-Net, combine the text embeddings and decoded feature maps to produce voxel-wise attention maps that guide segmentation. The decoder mirrors the encoder’s structure, with upsampling operations progressively reconstructing the segmentation map. The model was trained end-to-end using supervised learning, with the pre-trained language model frozen during training to minimize computational overhead and memory usage.

\subsection{Model Training}
We used the AdamW optimizer, with a starting learning rate of 1e-5 with a cosine schedule and a warmup of 5 epochs. An equal-weight compound loss function of Dice loss and cross-entropy was used. The PET and CT images were scaled to have an intensity between 0 and 1. We used slight augmentations of RandAffine, RandGaussianSmooth, and RandGaussianNoise, which randomly rotate, scale, smooth, and add noise during the training. The models were trained for 250 epochs on an NVIDIA A100 GPU, with the full training dataset requiring an average epoch training time of 3.6 hours and 38 days to fully train. 

\subsection{Evaluation}
The data was split at the patient level into a training set (9225/11356), a validation set (960/11356), and a testing set (1171/11356). To assess the accuracy of the automatic labeling pipeline, a board-certified radiologist manually reviewed 251 of the 1171 samples in the held-out testing set and scored the accuracy of the labels. The radiologist was given the extracted axial slice and SUVmax to ensure the correct lesion was labeled. This manually reviewed subset served as our final ground truth for model comparisons. The 1,171 algorithmically labeled holdout test set was used to analyze the model's performance across lesion size, tracer type, and uptake dependence, providing a greater number of samples for data analysis. This larger test set was not reviewed by a physician due to its size and the associated time constraints.

\subsection{Comparator Methods}

We compared our model with LLMSeg \cite{oh_llm-driven_2024}, an open-source 3D vision-language model. LLMSeg employs residual CNN encoders and Llama2-7B-chat as its language encoder, alongside an attention mechanism adapted from the Segment Anything Model (SAM) \cite{kirillov_segment_2023}. This attention mechanism compresses text representations into a few token embeddings, reducing the sequence length. As an additional comparison, we replaced LLMSeg’s attention module with the attention module from ConTEXTual Net 3D, allowing us to assess the efficacy of using compressed text representations on longer sequences.
In addition, we used ConTEXTual Net 2D to predict the lesion location in the coronal and sagittal planes with the coronal and sagittal maximum intensity projections (MIPs). The coronal and sagittal prediction masks were then extruded along their compressed dimension to create two 3D volumes, each representing the predicted spatial extent of the lesion in its respective plane. The final reconstructed 3D prediction was obtained by taking the intersection of these two volumes. 
Finally, as a human-level benchmark, we assigned the same visual grounding task to two practicing nuclear medicine physicians, which provided a benchmark for comparison. The physicians were given 101 images and referring expression pairs, which were a subset of the 251 images and sentences that had previously undergone manual review. Notably, the physicians were not provided with the SUVmax or the axial slice information when performing the task.

\subsection{Sensitivity Studies}

We explored the impact of using different LLMs within the ConTEXTual Net 3D architecture to evaluate the trade-off between model specialization and size. Due to long training times and large dataset size, we used 25\% of the data for the sensitivity studies to reduce computational requirements. The models tested included RoBERTa-Large \cite{liu_roberta_2019}, RadBERT \cite{yan_radbert_2022}, BERT \cite{devlin_bert_2019}, and LLama3.1 \cite{grattafiori_llama_2024}.  We also conducted experiments where the model did not have access to the text, isolating the influence of the vision component on performance. Additionally, we tested using the token embeddings directly after the first embedding layer, bypassing the remaining language model, to determine the extent to which the actual language modeling of tokens contributes to performance compared to the mere presence of relevant tokens. 
To assess the effect of training dataset size on model performance, we conducted a subsampling experiment. The model was trained on progressively smaller subsets of the full dataset, using proportions of 100\%, 50\%, 25\%, and 10\%. Performance was then evaluated on the physician-annotated test set to determine how reduced training data impacts the model's accuracy.

The full holdout test results were analyzed across several dimensions, including lesion size, radiotracer type, and SUVmax value. Lesion sizes were categorized as "small" (volume < 1 ml), "medium" (1–4 ml), and "large" (volume > 4 ml). Radiotracer distribution included four types: FDG, DOTATATE, [18F]Fluciclovine, and [18F]DCFPyL, with FDG accounting for 91\% of the test set (1,066 out of 1,171). Lesions were further grouped based on their SUVmax values into three ranges: 2.5–5.0 SUV, 5.0–10.0 SUV, and >10.0 SUV.

\subsection{Evaluation Metrics and Statistical Analysis}

We tracked four metrics for each experiment: matching SUVmax, any overlap, above 0.5 Dice, and average Dice score. The first three metrics were computed as F1 scores, where a lesion is considered a true positive if the prediction overlaps the target lesion and meets the corresponding criteria. For “matching SUVmax”, true positives occurred when the segmented lesion overlaps the target lesion, and the segmented SUVmax is within ±0.1 of the target SUVmax. For the “any overlap” metric, a true positive is identified if any voxel of the predicted lesion overlaps with the target lesion. The “above 0.5 Dice” metric requires the lesion-level Dice coefficient to exceed 0.5 for a lesion to be marked as a true positive. Finally, the average Dice score provides the mean Dice coefficient across all lesions, offering an overall measure of segmentation accuracy. A false positive is defined as any predicted lesion that does not satisfy all conditions, while a false negative occurs when a lesion is missed. Since our data does not include true negatives, the evaluation focuses solely on lesion-level metrics. To account for variability in the results, we use bootstrap random sampling with replacement for 10,000 iterations, and report 95\% confidence intervals for each metric. 

\section{Results}

Our weak labeling pipeline was found to generate a contour in the correct location 98\% (246/251) of the time, with boundary adjustments needed in 7.5\% (19/251) of the cases, as assessed by a physician. 

\subsection{Model Perforamnce}

\begin{figure*}[!h]
\centerline{\includegraphics[width=0.9\textwidth]{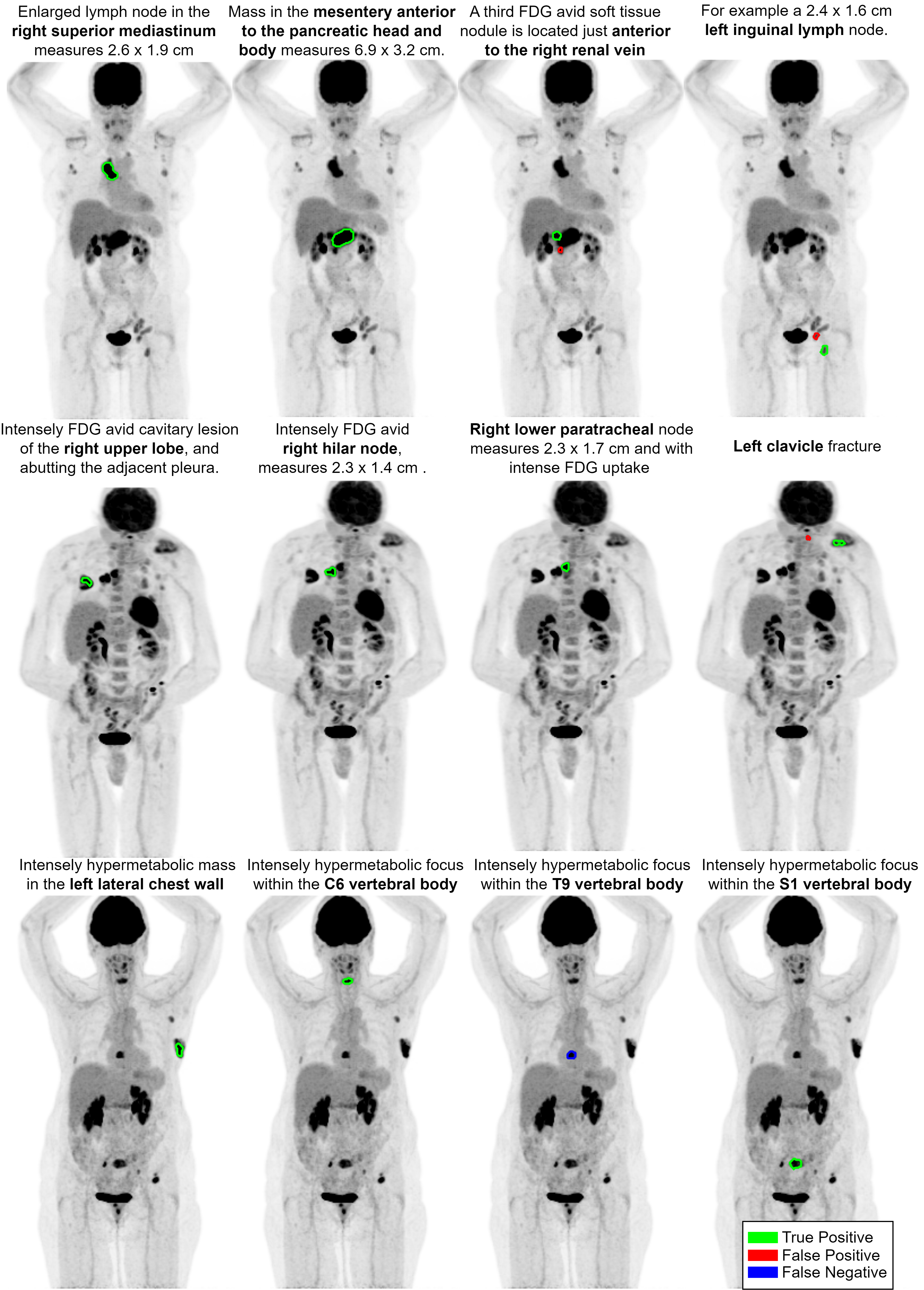}}
\caption{Example images and descriptions overlaid with the model outputs. True positives are shown in green, false positives are shown in red, and false negatives are shown in blue.}
\label{Examples}

\end{figure*}

ConTEXTual Net 3D achieved an F1 score of 0.80 on the physician-labeled test set, significantly outperforming LLMSeg (F1 = 0.22) and the 2.5D model (F1 = 0.53) (P < 0.05) as seen in Table \ref{table:comparator}. Examples of ConTEXTual Net 3D’s outputs can be seen in Figure \ref{Examples}. However, the model’s performance was lower than that of two board-certified radiologists (F1 = 0.94 and 0.91). ConTEXTual Net 3D surpassed the radiologists in average Dice score and lesion-level Dice > 0.5, likely due to the model’s exposure to thousands of training examples that were labeled with the same segmentation method.

\begin{table}[H]
\centering
\caption{Comparator Methods}
\label{table:comparator}
\resizebox{\columnwidth}{!}{%
\begin{tabular}{ccccc}
\toprule
Model Type & Matching SUVmax (F1)  & Any Overlap (F1)  &  Dice > 0.5 (F1) & Dice\\
\hline
\midrule
3D Comparisons & &\\
\cmidrule[1pt]{1-1}
ConTEXTual Net 3D & 0.798 [0.749, 0.843]  & 0.822  [0.774, 0.866] & 0.760 [0.709, 0.810] & 0.608  [0.566, 0.647] \\
ConTEXTual Net 2.5D  & 0.526 [0.462, 0.590] & 0.583 [0.520, 0.644] & 0.480 [0.414, 0.543] & 0.303 [0.258, 0.349]\\
LLMSeg  & 0.215 [0.164, 0.267] & 0.247 [0.193, 0.303] & 0.180 [0.133, 0.230] & 0.137 [0.103, 0.172]\\
LLMSeg (ConTEXTual net attention)  & 0.507 [0.446, 0.569] & 0.569 [0.507, 0.630] & 0.507 [0.444, 0.568] & 0.388 [0.343, 0.434] \\
\midrule
Human Performance & & \\
\cmidrule[1pt]{1-1}
Physician 1 & 0.912 [0.853, 0.961] &  0.941 [0.892, 0.980] & 0.451 [0.353, 0.549] & 0.407 [0.359, 0.455]\\ 
Physician 2  &  0.942  [0.886, 0.990] & 0.980 [0.951, 1.000] & 0.667 [0.569, 0.755] & 0.539 [0.492, 0.583]\\ 
\bottomrule
\end{tabular}
}
\end{table}

\subsection{Training Data Efficiency}

\begin{figure*}[!h]
\centerline{\includegraphics[width=0.9\textwidth]{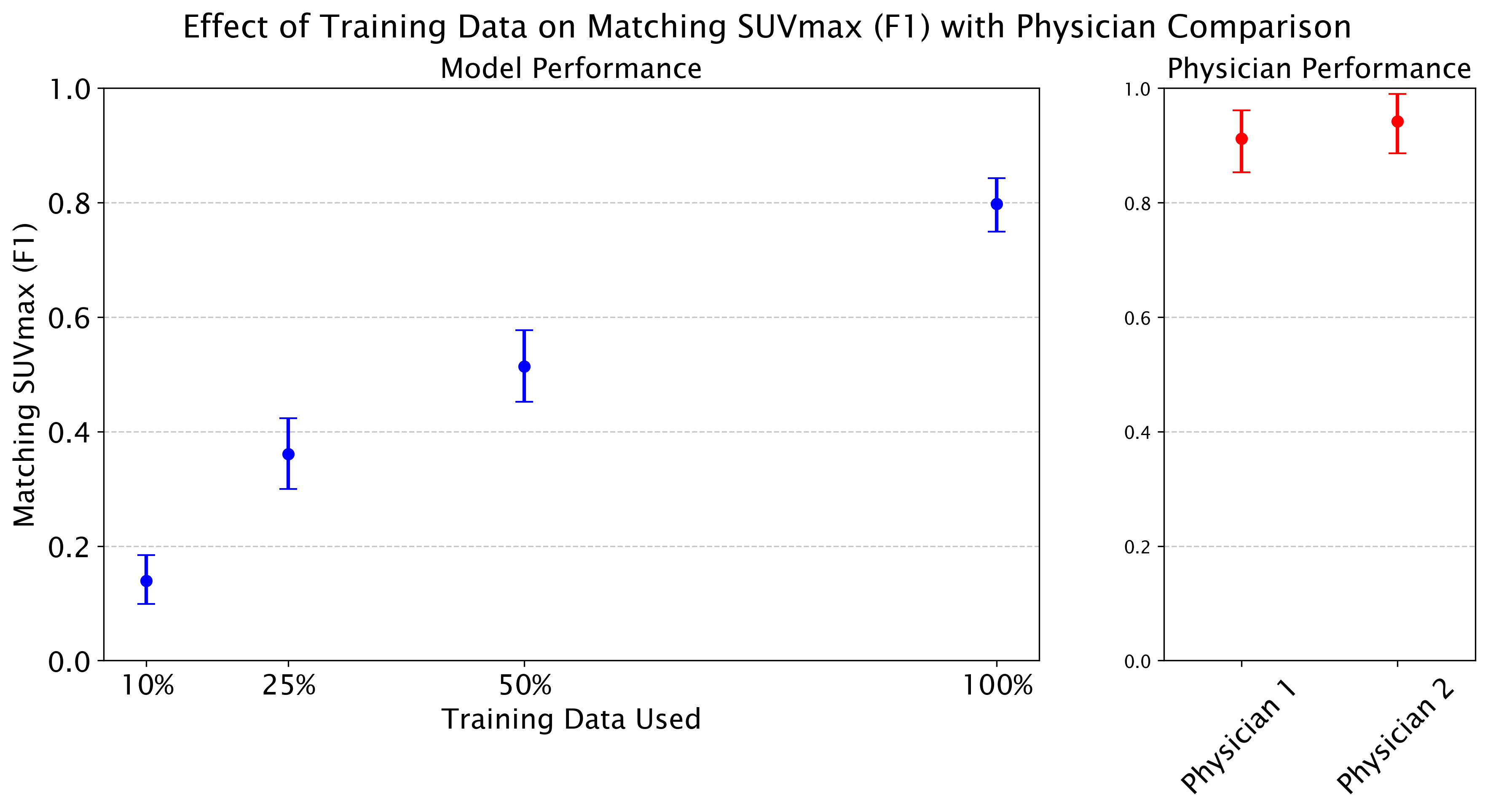}}
\caption{Performance of ConTEXTual Net 3D across varying training data sizes compared to physician benchmarks, with confidence intervals.}
\label{perforamnce_plot}

\end{figure*}

Figure \ref{perforamnce_plot} illustrates the performance of ConTEXTual Net 3D as a function of training data size, alongside a comparison to physician performance. The model’s F1 score consistently improved with increased training data, reaching a peak of 0.80 [95\% CI: 0.75, 0.84] when trained on the full dataset. Tabular results for 10\%, 25\%, and 50\% training data are provided in the appendix Table \ref{table:data_scaling}.

\subsection{Sensitivity Studies}

The performance of ConTEXTual Net 3D was evaluated when using different language models, which are shown in Table \ref{table:language_model}. RadBERT demonstrated the highest F1 score (0.36), suggesting an advantage for domain-specific language models for this task. Generalist models, including BERT (F1 = 0.31), RoBERTa-Large (F1 = 0.27), and LLama 3.1 8B (F1 = 0.30), performed comparably. Models without language models, such as ConTEXTual Net 3D (No report, F1 = 0.13) and ConTEXTual Net 3D Embeddings without LLM (F1 = 0.19), performed substantially worse, indicating that the task is not suitable for vision-only models.

\begin{table}[H]
\centering
\caption{Large Language Model Study}
\label{table:language_model}
\resizebox{\columnwidth}{!}{%
\begin{tabular}{ccccc}
\toprule
Model Type & Matching SUVmax (F1)  & Any Overlap (F1)  &  Dice > 0.5 (F1) & Dice\\
\hline
\midrule

\midrule
Language Models Trained on 25\% of Data & & & & \\
\cmidrule[1pt]{1-1}

ConTEXTual Net 3D (RadBERT) & 0.361 [0.300, 0.423] & 0.381 [0.320, 0.443] & 0.341 [0.281, 0.403]  & 0.240 [0.197, 0.284]\\
ConTEXTual Net 3D (BERT) & 0.315 [0.258, 0.374] & 0.331 [0.272, 0.390] & 0.272 [0.216, 0.329]  & 0.219 [0.178, 0.261]\\
ConTEXTual Net 3D (RoBERTa-Large) & 0.271 [0.216, 0.326] &  0.289 [0.232, 0.344] & 0.244 [0.191, 0.298] & 0.187 [0.148, 0.226]\\
ConTEXTual Net 3D (Llama3.1)  & 0.296 [0.237, 0.354] & 0.316 [0.256, 0.377] & 0.271 [0.213, 0.330] &  0.184 [0.146, 0.224]\\
ConTEXTual Net 3D (Embeddings No LLM) & 0.186 [0.138, 0.236] &  0.191 [0.142, 0.241] & 0.174 [0.128, 0.222] & 0.130 [0.096, 0.165]\\
ConTEXTual Net 3D (No Report) &  0.132 [0.092, 0.175] & 0.136 [0.096, 0.180]  & 0.120 [0.080, 0.161] & 0.092 [0.063, 0.122] \\
\bottomrule
\end{tabular}
}
\end{table}

\subsection{Data Analysis}

ConTEXTual Net 3D performed best on FDG PET/CT exams (F1=0.78) and [18F] DCFPyL (F1=0.75), but performance dropped for both DOTATE (F1=0.58) and [18F]Fluciclovine (F1=0.66). The model showed similar performance across lesion sizes but suffered a drop in performance on low uptake lesions between 2.5-5.0 SUV (F1=0.70) when compared to 5.0-10.0 SUV(F1=0.80) and above 10.0 SUV (F1=0.77) as shown in Table \ref{table:data_analysis}.

\begin{table}[H]
\centering
\caption{Data Analysis}
\label{table:data_analysis}
\resizebox{\columnwidth}{!}{%
\begin{tabular}{ccccc}
\toprule
Model Type & Matching SUVmax (F1)  & Any Overlap (F1)  &  Dice > 0.5 (F1) & Dice\\
\hline
\midrule
Lesion Size Analysis & & & & \\
\cmidrule(lr){1-1}
lesions < 1 Milliliter (n = 322)   & 0.775 [0.728, 0.820] & 0.751 [0.702, 0.798] & 0.768 [0.720, 0.813] & 0.569 [0.532, 0.605]\\
lesions 1 - 4 Milliliters (n = 461)  & 0.773 [0.733, 0.811] & 0.732 [0.691, 0.772] & 0.760 [0.721, 0.798] & 0.588 [0.555, 0.620]\\
lesions > 4 Milliliters (n = 388) & 0.739 [0.693, 0.782] &  0.696 [0.648, 0.741] & 0.709 [0.663, 0.753] & 0.558 [0.520, 0.594]\\
\midrule
Tracer Type Analysis & & & & \\
\cmidrule(lr){1-1}
FDG (n=1066) & 0.775 [0.750, 0.800] & 0.742 [0.716, 0.768] & 0.759 [0.733, 0.783] & 0.587 [0.566, 0.608]\\
DOTATATE (n=58) & 0.579 [0.448, 0.702] & 0.491 [0.362, 0.621] & 0.544 [0.414, 0.672] & 0.383 [0.287, 0.481] \\
{}[18F]Fluciclovine (n=30) & 0.655 [0.481, 0.814] & 0.586 [0.407, 0.759] & 0.655 [0.481, 0.814] & 0.465 [0.329, 0.600]\\
{}[18F]PYL (n=17) & 0.750 [0.533, 0.938] & 0.688 [0.452, 0.882] & 0.750 [0.533, 0.938] & 0.523 [0.350, 0.683]\\
\midrule
Uptake Analysis & & & & \\
\cmidrule(lr){1-1}
2.5 < SUVmax < 5 (n = 343) & 0.701 [0.652, 0.748] & 0.670 [0.620, 0.718] & 0.686 [0.637, 0.733] & 0.505 [0.465, 0.543]\\
5 < SUVmax < 10 (n = 506)  & 0.800 [0.728, 0.820] & 0.773 [0.736, 0.808] & 0.785 [0.750, 0.821] & 0.619 [0.590, 0.648]\\
10 < SUVmax (n = 322)      & 0.765 [0.718, 0.811] & 0.707 [0.658, 0.757] & 0.743 [0.696, 0.789] & 0.573 [0.534, 0.612]\\
\bottomrule
\end{tabular}
}
\end{table}

\section{Discussion}

We developed an automated pipeline that used the noted SUVmax and axial slice to create labels for lesions paired with their referring expression. This pipeline was found to be highly accurate and capable of providing a dataset suitable for 3D visual grounding in the medical domain. The dataset was then used to train a vision-language model, ConTEXTual Net 3D, which shows promising capability at locating the described disease but did not perform as well as practicing physicians. Our work shows the feasibility of vision-language modeling in PET/CT using weak labeling.       

Some cases can be challenging due to the inherent ambiguity in textual descriptions. For example, in {Firgure 4}, the first row, last column, a description of a "left inguinal node" leads the model to segment both left inguinal nodes. While either segmentation could be considered correct, our labeling process assigns only one node with a correct SUVmax and axial slice, leaving the other unlabeled. This ambiguity reflects why radiologists include specific details like SUVmax and axial slice numbers in their reports—to disambiguate similar findings. Addressing this type of ambiguity is likely to remain a core challenge for visual grounding in nuclear medicine.

Our model significantly outperformed comparator methods. The longer textual descriptions commonly found in nuclear medicine reports introduce a unique challenge for visual grounding. LLMSeg struggled with these, whereas its performance improved when its attention module was replaced with that of ConTEXTual Net 3D, underscoring the limitations of approaches that compress language representations. While such methods reduce computational demands, they appear suboptimal for long-sequence visual grounding tasks. Our model also outperformed conTEXTual Net 2.5D, which operates on 2D sagittal and coronal MIPs. This highlights the importance of developing 3D methods for PET/CT rather than compressing the visual information.   

The model's performance surprisingly showed minimal dependence on lesion size. Smaller lesions were often well-described and disambiguated, whereas larger lesions, commonly found in conglomerates or bulky masses, could pose challenges for precise boundary definition and SUVmax estimation despite being visually apparent. Performance was highest on FDG, which dominates the dataset (>90\%), while lower performance on DOTATATE and [18F]Fluciclovine aligns with their lower uptake levels and, in the latter case, a more dispersed nature. Similarly, lower uptake lesions (SUVmax 2.5–5.0) were more challenging, likely due to their subtlety and the increased risk of segmenting nearby lesions. The model performed comparably on medium (SUVmax 5.0–10.0) and high uptake (SUVmax >10.0) lesions.

ConTEXTual Net 3D’s performance fell short of the nuclear medicine physicians’ performances. Our substudy on the impact of sample size indicated that scaling up the dataset to include more exams, which would require multiple centers, could help close the gap between physicians and the vision-language model. 

There are several limitations to this work, a key limitation of this study is that the training data consists solely of sentences from radiology reports where the radiologist explicitly noted an SUVmax value and an axial slice. This may introduce selection bias, as lesion types that are less frequently documented with SUVmax and axial slices are underrepresented or entirely absent. Furthermore, sentences containing multiple axial slice numbers were excluded from the process, which could limit the model’s ability to handle descriptions involving multiple lesions. Lesions with SUVmax values below 2.5 were also excluded due to the cutoff we applied, potentially overlooking lower-uptake abnormalities. This cut-off was required due to the high number of connected components found at low uptake thresholds that would intersect the specified plane, which introduced many false positives. Additionally, this work does not address organ-level segmentations, unless accompanied by an SUVmax value and axial slice— but these details are often absent in organ-level observations. Finally, this work focuses on a single center, so it is unclear how well this model will generalize given the heterogeneity of reporting styles and image quality found at different centers. 

\section{Conclusion}

This work demonstrates the potential of vision-language models to bridge radiology reports and PET/CT imaging for accurate lesion localization and segmentation. Our novel weak labeling pipeline achieved high accuracy in creating annotated PET/CT image-text pairs, enabling the development of 3D visual grounding models. While ConTEXTual Net 3D significantly outperformed other models, it falls short of the performance of nuclear medicine physicians, highlighting the need for even larger datasets to close this gap.



\bibliography{references}  
\newpage

\appendix

\counterwithin{figure}{section}
\counterwithin{table}{section}
\renewcommand{\thesection}{\Alph{section}}
\renewcommand{\thefigure}{\thesection\arabic{figure}}
\renewcommand{\thetable}{\thesection\arabic{table}}

\section{Supplementary Tables and Figures}

\begin{table}[H]
\centering
\caption{Post-processing Techniques}
\label{table:post_processing}
\resizebox{\columnwidth}{!}{
\begin{tabular}{ccccc}
\toprule
Model Type & Matching SUVmax (F1)  & Any Overlap (F1)  &  Dice > 0.5 (F1) & Dice\\
\hline
\midrule
Post-Processing Techniques & & & & \\
\cmidrule[1pt]{1-1}
Lesion with highest average logits  & 0.798 [0.749, 0.843]   & 0.822 [0.774, 0.866] & 0.760 [0.709, 0.810] & 0.608 [0.566, 0.647] \\
Lesion with highest logit          & 0.773 [0.720, 0.823] & 0.796 [0.745, 0.843] & 0.764 [0.710, 0.815] &  0.602 [0.559, 0.644] \\
No post-processing  & 0.713 [0.663, 0.763] & 0.752 [0.706, 0.797] & 0.764 [0.710, 0.815] & 0.615 [0.575, 0.653]\\
\bottomrule
\end{tabular}
}
\end{table}

\begin{table}[H]
\centering
\caption{Data Scaling Analysis}
\label{table:data_scaling}
\resizebox{\columnwidth}{!}{%
\begin{tabular}{ccccc}
\toprule
Model Type & Matching SUVmax (F1)  & Any Overlap (F1)  &  Dice > 0.5 (F1) & Dice\\
\hline
\midrule
Varying Training Data Quantity & & & &\\
Performance Scaling on Data Size & & & &\\
\cmidrule[1pt]{1-1}
\cmidrule[1pt]{1-1} 
100\% (9225/9225) Training Data       & 0.798 [0.749, 0.843] & 0.822 [0.774, 0.866] & 0.760 [0.709, 0.810] & 0.608 [0.566, 0.647] \\
50\% (4613/9225) Training Data        & 0.514 [0.452, 0.577] & 0.535 [0.474, 0.599] & 0.505 [0.443, 0.569] & 0.370 [0.323, 0.419]\\
25\% (2308/9225) Training Data       & 0.361 [0.300, 0.423] & 0.381 [0.320, 0.443] & 0.341 [0.281, 0.403] & 0.240 [0.197, 0.284]\\
10\% (921/9225) Training Data        & 0.139 [0.099, 0.184] & 0.140 [0.099, 0.184] & 0.104 [0.068, 0.143] & 0.089 [0.062, 0.118]\\
\bottomrule
\end{tabular}
}
\end{table}

\begin{figure*}[!h]
\centerline{\includegraphics[width=0.9\textwidth]{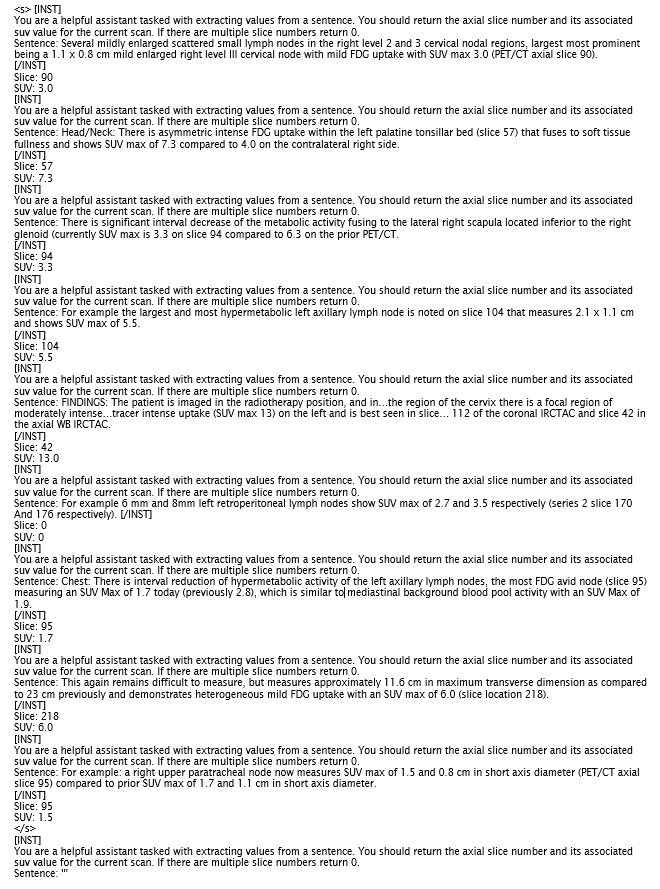}}
\caption{Multi-shot in context prompt used to extract SUVmax and axial slice numbers for the present scan.}
\label{prompt}
\end{figure*} 

\end{document}